\documentclass[runningheads]{llncs}
\usepackage{graphicx}
\usepackage{amsfonts}
\usepackage{amsmath}
\usepackage{booktabs}

\begin{document}
\title{Dynamic Depth-Supervised NeRF for Multi-View RGB-D Operating Room Videos}
\titlerunning{Dynamic DS-NeRF for Multi-View RGB-D OR Videos}
%
\author{Beerend G.A. Gerats\inst{1,3}\orcidID{0000-0003-2570-1834} \and
Jelmer M. Wolterink\inst{2}\orcidID{0000-0001-5505-475X} \and
Ivo A.M.J. Broeders\inst{1,3}\orcidID{0000-0001-7524-9263}}
\authorrunning{B.G.A. Gerats et al.}
%
\institute{
    Centre for Artificial Intelligence, Meander Medisch Centrum, Amersfoort, The Netherlands \email{\{initials.surname\}@meandermc.nl}\and
    Department of Applied Mathematics \& Technical Medical Center, University of Twente, Enschede, The Netherlands \and
    Robotics and Mechatronics, University of Twente, Enschede, The Netherlands
}
\maketitle              
\begin{abstract}
The operating room (OR) is an environment of interest for the development of sensing systems, enabling the detection of people, objects, and their semantic relations. Due to frequent occlusions in the OR, these systems often rely on input from multiple cameras. While increasing the number of cameras generally increases algorithm performance, there are hard limitations to the number and locations of cameras in the OR. Neural Radiance Fields (NeRF) can be used to render synthetic views from arbitrary camera positions, virtually enlarging the number of cameras in the dataset. In this work, we explore the use of NeRF for view synthesis of dynamic scenes in the OR, and we show that regularisation with depth supervision from RGB-D sensor data results in higher image quality. We optimise a dynamic depth-supervised NeRF with up to six synchronised cameras that capture the surgical field in five distinct phases before and during a knee replacement surgery. We qualitatively inspect views rendered by a virtual camera that moves 180 degrees around the surgical field at differing time values. Quantitatively, we evaluate view synthesis from an unseen camera position in terms of PSNR, SSIM and LPIPS for the colour channels and in MAE and error percentage for the estimated depth. We find that NeRFs can be used to generate geometrically consistent views, also from interpolated camera positions and at interpolated time intervals. Views are generated from an unseen camera pose with an average PSNR of 18.2 and a depth estimation error of 2.0\%. Our results show the potential of a dynamic NeRF for view synthesis in the OR and stress the relevance of depth supervision in a clinical setting.

\end{abstract}
\section{Introduction}
The operating room (OR) is an environment of interest for the development of sensing systems, with tasks ranging from person detection and human pose estimation \cite{hansen2019} to domain modeling and role identification \cite{ozsoy20224dor}. These sensing systems could automatically register adverse events and distractions \cite{goldenberg2017using} or monitor X-ray radiation exposure \cite{padoy2019} to enhance patient and staff safety. Due to frequent occlusions by large devices, hanging monitors and a crowded space, detection systems often rely on video input from multiple cameras. Typically, the underlying algorithms perform better when the number of cameras is increased \cite{gerats20223d}. However, there are limitations to the number and locations of cameras in the OR due to sterility concerns and limited available space.

Neural Radiance Fields (NeRF) \cite{mildenhall2020nerf} is a powerful technology for the reconstruction of a 3D scene from a set of images that capture the scene from various camera positions. Although the introduction of this technology has caused an explosion of interest in the field of computer vision, with many follow-up studies \cite{xie2022neural}, NeRF-based methods for clinical use remain largely unexplored \cite{wang2022neural}. The use of NeRF could help to overcome the limited availability of camera placement in the OR, by virtually increasing the number of cameras with view synthesis from new camera locations. Subsequently, renders of OR scenes from arbitrary camera angles could be used to improve detection algorithms.

In this paper, we explore the use of NeRF for view synthesis of dynamic scenes in the OR and we show that regularisation with depth supervision \cite{deng2022depth} increases the render quality and reduces the need for many camera positions. We find that a depth-supervised NeRF optimised with only six synchronised camera views is able to generate images of the surgical intervention from a range of camera angles. In contrast to existing depth supervision in NeRF, we directly optimise our model using measured RGB-D sensor data instead of estimated depth from a structure from motion (SfM) algorithm. Additionally, we extend the method with a notion of scene dynamics by optimising with an extra time variable enabling the reconstruction of the surgical scene at specific time intervals. We show how the dynamic NeRF could be used for region-of-interest localisation through unsupervised segmentation of objects and people in the OR.

\section{Related work}
\subsection{Neural Radiance Fields}
NeRF is a method for volume rendering, based upon the \textit{implicit representation} of a 3D scene in the weights of a neural network $F_\Theta$ \cite{mildenhall2020nerf}. This network is generally a standard multi-layer perceptron (MLP) that takes a 5D vector $(x, y, z, \theta, \phi)$ as input and that outputs a 4D vector $(RGB, \sigma)$. An input vector consists of a 3D location $(x, y, z)$ in the captured scene and an orientation $(\theta, \phi)$ from which this location is viewed. $F_\Theta$ returns for each vector a colour $RGB$ and a volume density $\sigma$. With this simple setup, NeRF can reconstruct images by casting a viewing ray from each pixel, sampling points along that ray, asking the MLP to find the colours and densities for these points and to sum over these results. In this way, it is possible to use a discrete set of sampled points in the 3D scene, while representing the scene in continuous form. Reconstructed images are compared with ground truth images that are taken from the same camera positions. The rendering function that sums over the colours and densities is differentiable such that the MLP can be optimised by stochastic gradient descent. The loss function is often a mean squared error between the colours of the rendered and the ground truth images.

\subsection{NeRF with Depth Priors}
Although NeRF has the ability to synthesise photo-realistic images from unseen camera perspectives, the method does not guarantee to capture 3D geometry accurately. This limitation becomes visible when rendering poorly textured areas that often occur in indoor scenes \cite{wei2021nerfingmvs}. It is likely that this problem will play a role when NeRF is applied to the OR.

Several solutions are proposed that involve regularisation with depth priors. Nerfing MVS \cite{wei2021nerfingmvs} provides a guided optimisation scheme, where points are sampled along a viewing ray only around depth values found earlier. A sparse set of depth values is found by applying the COLMAP SfM algorithm \cite{schonberger2016structure} on multi-view images. The sparse sets are used to train a depth completion network that provides full sets of depth values. A more common approach is to use a depth loss that enforces NeRF to represent a large amount of volume density around ground truth depth values \cite{rematas2022urban}, or that enforces NeRF to terminate rays close to depth observations, provided by a depth completion network \cite{roessle2022depthpriorsnerf}, the COLMAP algorithm or RGB-D data \cite{deng2022depth}.

\subsection{NeRF for Dynamic Scenes}
While the original NeRF has been developed to represent static scenes, several adaptations have been proposed to give NeRF a sense of scene dynamics. D-NeRF \cite{pumarola2021d} uses an additional deformation network that models point translation to a canonical configuration. With this network, it is possible to render 3D objects that change in shape over time. However, it is less suitable for the clinical setting, where people and objects can appear and disappear in the reconstructed scene. Li \textit{et al.} \cite{li2022neural} add temporal latent codes to the input vectors for video synthesis. To speed up training, they propose hierarchical frame selection and importance sampling of camera rays.

\subsection{NeRF for Clinical Interventions}
Although NeRF has inspired several works in medical image computing \cite{wolterink2022implicit}\cite{sun2021coil}, it has not been widely adopted in the field of computer-assisted interventions. To the best of our knowledge, the technology has thus far only been used for 3D reconstruction of soft tissues in robotic surgery videos \cite{wang2022neural}. Because the laparoscopic view is a single view, their method uses time-dependent modelling of neural radiance and displacement fields, based on D-NeRF. Using the stereoscopic camera of the surgical robot, the method finds depth maps along the coloured image frames. The depth information is used to constrain NeRF optimisation with an additional loss function.

\section{Methods}

\subsection{Dynamic Depth-Supervised NeRF}
We use the depth-supervised NeRF (DS-NeRF) by Deng \textit{et al.} \cite{deng2022depth} for building 3D reconstructions of OR scenes. This method regularises the training with an additional depth loss such that a model can be optimised with relatively few camera positions. The key idea in DS-NeRF is that most viewing rays terminate at the closest surface, which is often opaque. Therefore, most volume density should be found close to the distance of this surface along the viewing ray. DS-NeRF enforces such a distribution of volume density by minimising the KL divergence between the volume density distribution $h_i(s)$ and a normal distribution around the ground truth depth $d_i$ of keypoint $x_i \in X$:

\begin{equation}
    \text{KL}[\mathbb{N}(d_i, \hat{\sigma}_i) \| h_i(s)],
\end{equation}

where $X$ is the set of all keypoints in an image for which the depth is known and $s$ is the far endpoint of the viewing ray. The variance $\hat{\sigma}_i$ is set to the uncertainty of the depth estimation for keypoint $x_i$. When depth is estimated with COLMAP, the uncertainty is calculated by re-projecting the keypoint to and from another camera position in which the keypoint is visible. In RGB-D data, however, the depth values are measurements rather than estimations. Therefore, we set $\hat{\sigma}_i = 1.0$ for all keypoints such that each depth value is weighted equally. We sample 100K depth values in each ground truth image at random positions, where pixels that have zero depth are never included.

The OR is a dynamic environment, where objects and people move around in a controlled manner. Training NeRF with a notion of these dynamics over time is a natural choice for the synthesis of video. Inspired by DyNeRF \cite{li2022neural}, we extend DS-NeRF to generate views dependent on a time variable $t$. Our Dynamic DS-NeRF takes a 6D vector $(x, y, z, \theta, \phi, t)$ as input and outputs the same 4D vector $(RGB, \sigma)$. In contrast to the other input variables, we do not encode the time variable. We concatenate the scalar to the positional embeddings of the locations.

\subsection{Dataset}
\begin{figure}[t!]%
    \centering
    \includegraphics[width=\textwidth]{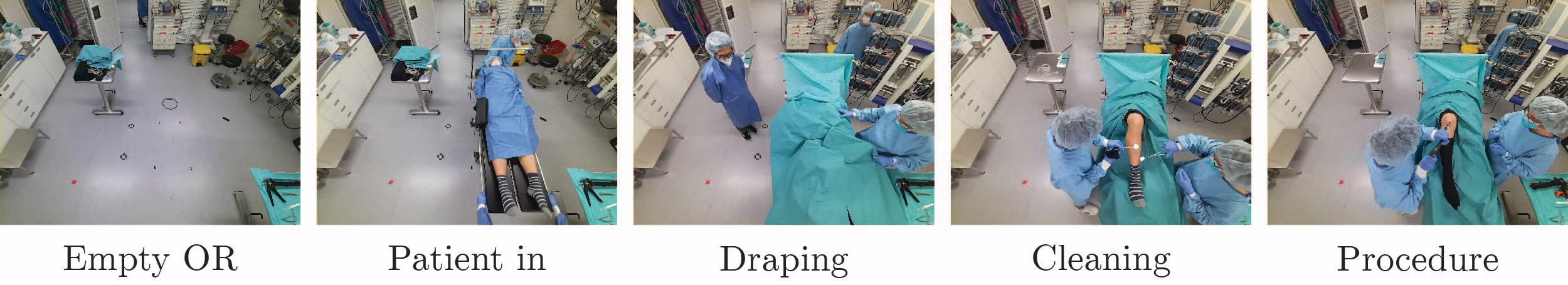}
    \caption{Five distinct phases during or before the surgery.}
    \label{fig:phase_examples}
\end{figure}

\begin{figure}[t!]%
    \centering
    \includegraphics[width=0.6\textwidth]{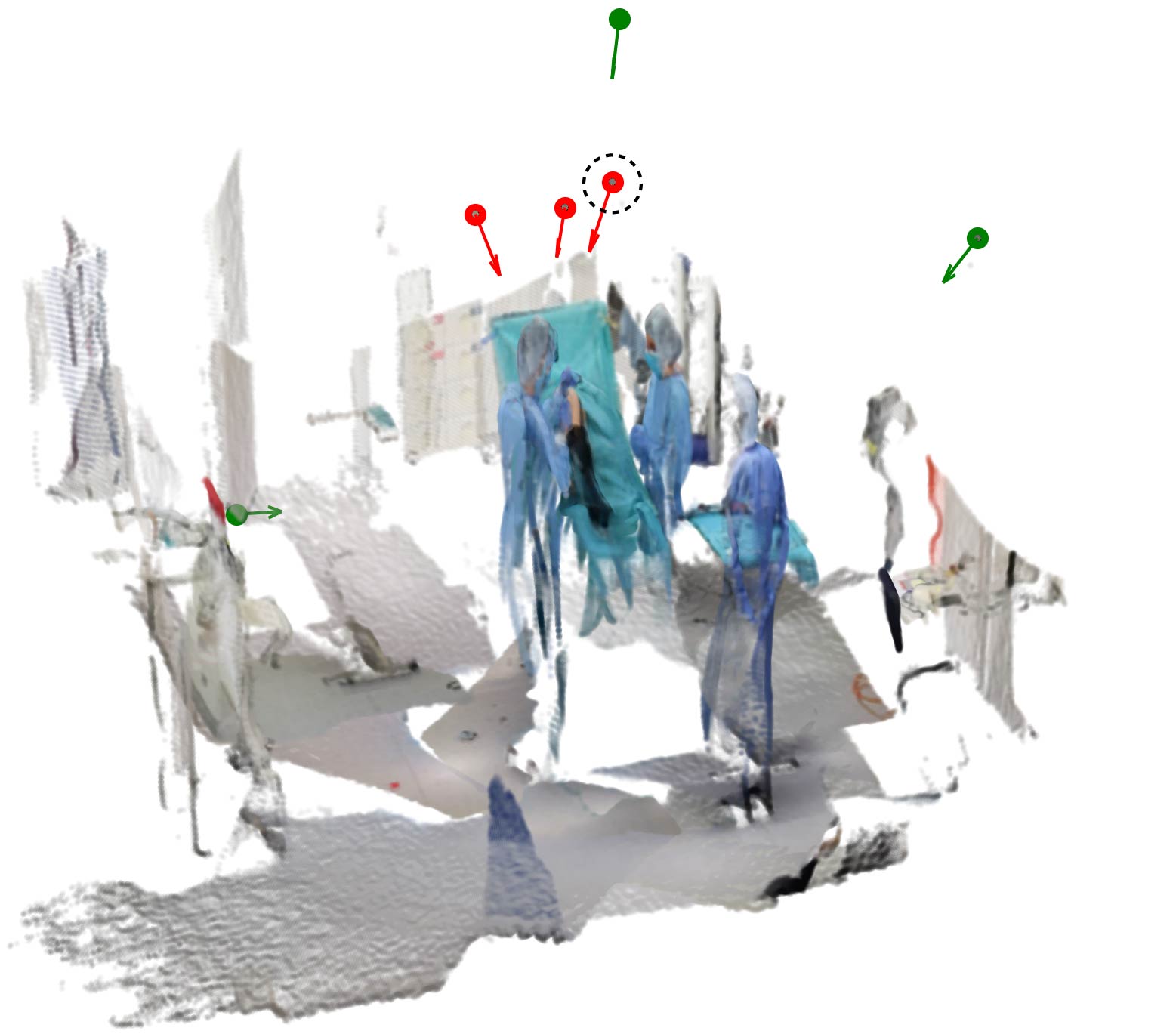}
    \caption{Locations of the RGB-D cameras are indicated by a red or green dot. The viewing angle is directed by the arrows. Red dots indicate three cameras located above the surgical field, whereas the green dots capture the OR from very different perspectives. The scene is a coloured point cloud formed by camera projection using depth values.}
    \label{fig:camera_poses}
\end{figure}

The 4D-OR dataset \cite{ozsoy20224dor} from the Technical University of Munich contains RGB-D images and camera poses from ten simulated knee replacement surgeries. We use this dataset to train the method in the reconstruction of five surgical phases (see Figure~\ref{fig:phase_examples}). In this way, we can evaluate the quality of synthesised views for different OR activities. The dataset contains synchronised images from six cameras that have fixed locations in the OR. Three cameras are located above the surgical field, each rotated approximately 90 degrees around the yaw axis (the red dots in Figure~\ref{fig:camera_poses}). Two of the three other cameras capture the OR from a wide perspective, while the sixth camera records from a position that is closer to the ground (the green dots).

\subsection{Experimental Setup}
First, we train Dynamic DS-NeRF with images from all six camera positions and qualitatively inspect a set of synthesised views. To obtain this set, we ask the algorithm to synthesise views from the three camera locations above the surgical field as well as from interpolated poses that together form a 180 degrees rotation around the surgical field. Second, we train the method with five camera positions and evaluate view synthesis from the remaining unseen sixth position (the circled camera pose in Figure~\ref{fig:camera_poses}). We evaluate the resulting images in terms of peak signal-to-noise ratio (PSNR), structural similarity index measure (SSIM) and learned perceptual image patch similarity (LPIPS). These metrics indicate image reconstruction quality in comparison to real images, where PSNR is derived from a pixel-wise mean squared error, SSIM indicates the inter-dependency of pixels that are spatially close \cite{wang2004image}, and LPIPS is based upon similarity in the deep representations of a convolutional neural network \cite{zhang2018unreasonable}. We evaluate the quality of depth perception in terms of mean absolute error (MAE) and in percentage of the ground truth depth value. Depth values equal to zero in the ground truth images are not evaluated, since these values are inaccurate measurements.

For each surgical phase, we optimise the method with $t=\{-2, -1, 0, 1, 2\}$ and evaluate the reconstructed scene at $t=0$ such that we can compare the results with non-dynamic methods. All models are trained to synthesise images of $512 \times 384$ pixels with the following hyperparameters: 768 neurons in each network layer, 4096 selected rays per batch, 192 points sampled per ray, 50K iterations and depth loss weighting factor $\lambda_D$ set to 0.1. This number of iterations takes around 20 hours of training time on a single NVIDIA A40 (48GB) GPU. At inference, the rendering of a single frame requires 22 seconds.

\section{Results}

\subsection{Qualitative Analysis}

\begin{figure}
    \centering
    \includegraphics[width=\textwidth]{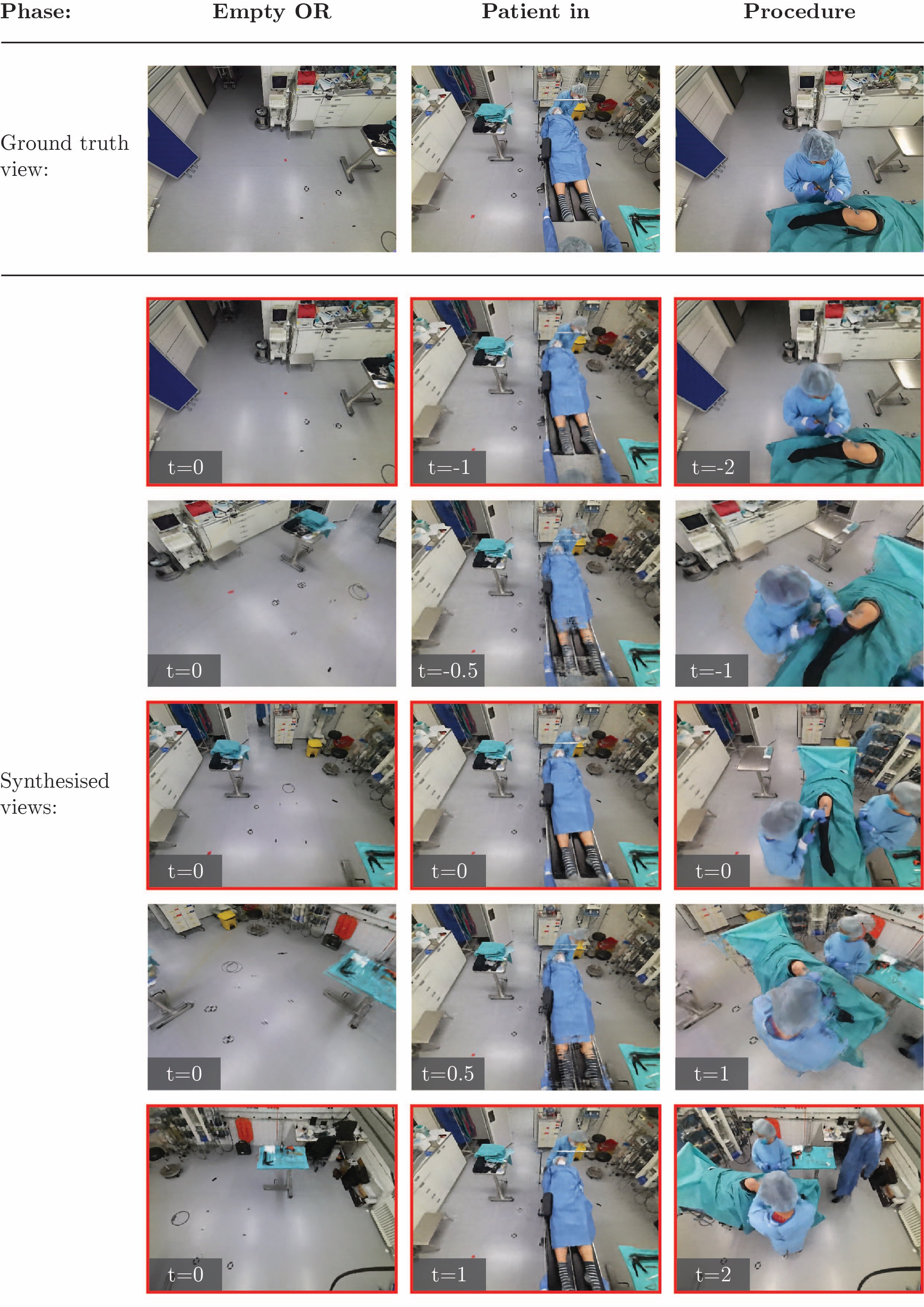}
    \caption{Dynamic DS-NeRF synthesised views for three phases in the OR at time intervals $t$. For ``empty OR'', the virtual camera rotates 180 degrees around the surgical field. For ``patient in'', the camera position is static, but the time intervals are interpolated. For ``procedure'', the images differ in camera location and time value. The top row displays the ground truth images for the starting camera pose. Views with a red border are generated from camera poses with which the algorithm is trained, corresponding to the red camera positions in Figure~\ref{fig:camera_poses}.}
    \label{fig:nerf_results}
\end{figure}

The synthesised views for three surgical phases are given in Figure~\ref{fig:nerf_results}. When comparing the top-row synthesised views with the ground truth images, it can be seen that Dynamic DS-NeRF is able to reconstruct the scenes independently of the surgical phase. Lighting conditions, such as reflections on the floor and shadows, are realistically rendered. However, the reconstructed scene looks smoothed with missing details, e.g., in the keyboard of the mobile monitor.

Synthesised views from the interpolated camera poses (images without red borders) correctly grasp the geometry of the scenes and find realistic lighting conditions. For the ``patient in'' in phase, we interpolate the time values at $t=-0.5$ and $t=0.5$. The moving patient bed is correctly positioned in these frames, at an interpolated location. However, the interpolated camera positions and time values induce a number of artefacts. First, various objects seem to be misaligned. For example, the tape on the floor and the instrument table are not always represented correctly. Second, fine-grained details, such as the surgical instruments in the hands of the left physician, are missing.

\subsection{Quantitative Analysis}
\begin{figure}[t!]%
    \centering
    \includegraphics[width=0.8\textwidth]{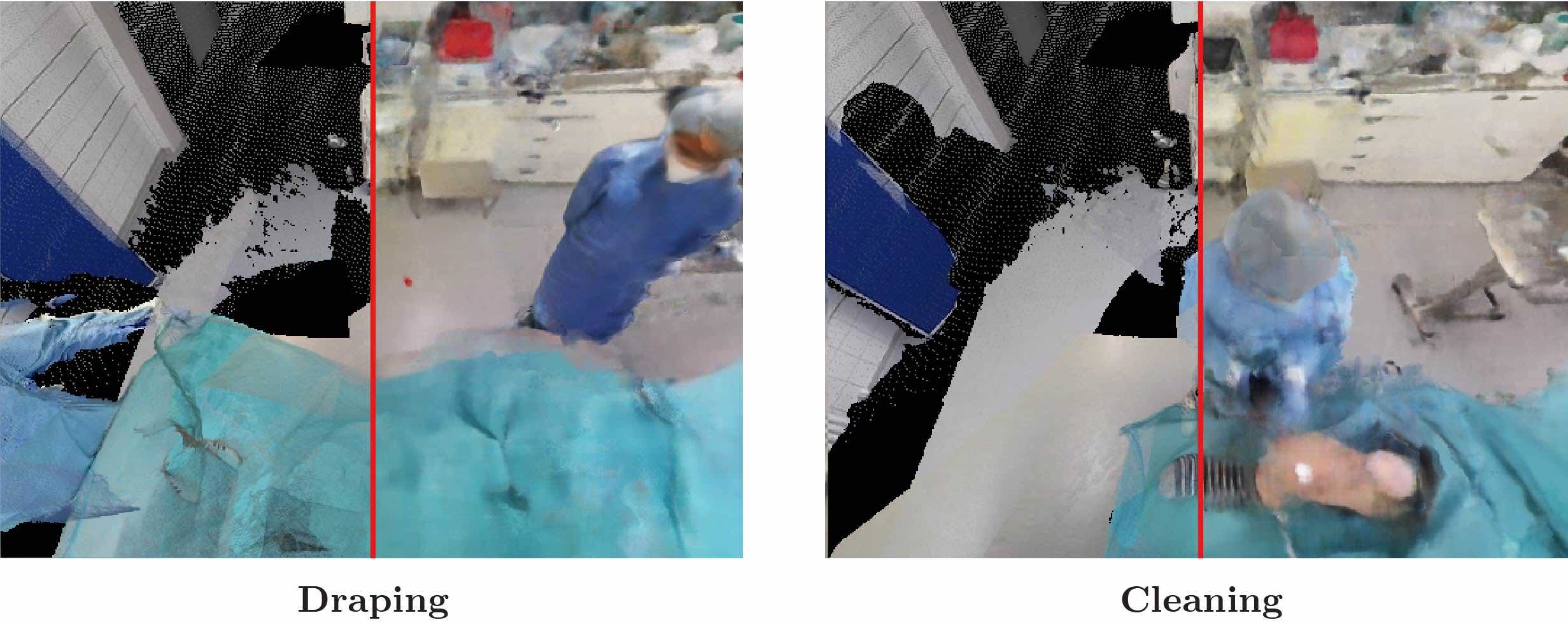}
    \caption{Left-right comparison between the pointcloud projection as baseline method (left) and output of Dynamic DS-NeRF (right) for an unseen camera pose in two surgical phases.}
    \label{fig:baseline_comparison}
\end{figure}

Results of the quantitative analysis can be found in Table~\ref{tab:results}. On average, Dynamic DS-NeRF is able to synthesise views from the unseen camera pose with 18.2 PSNR, 0.61 SIMM and 0.41 LPIPS. The image quality of the rendered views on the 4D-OR dataset differs per phase and ranges in PSNR from 17.5 for ``patient in'' to 19.6 for ``empty OR''. This shows that the image quality is dependent on the complexity of the surgical scene. Table~\ref{tab:comparison_results} provides a comparison with a naive baseline method for view synthesis that uses the projection of pointclouds to reconstruct the images. In this method, the five training images are projected into a single pointcloud, which is reprojected to the sixth camera position. Figure~\ref{fig:baseline_comparison} shows a left-right comparison between the baseline images and Dynamic DS-NeRF for two surgical phases. It can be seen that our method is able to generate proper colours for locations where the pointcloud has no presence. This results in more realistic-looking images and better image quality scores.

\begin{table}[t!]
\begin{center}
\begin{minipage}{\textwidth}
\caption{Evaluation metrics comparing Dynamic DS-NeRF synthesised views with unseen ground truth RGB-D images.}\label{tab:results}
\begin{tabular*}{\textwidth}{@{\extracolsep{\fill}}lccccc@{\extracolsep{\fill}}}
    \toprule
    & \multicolumn{3}{@{}c@{}}{Colour Image} & \multicolumn{2}{@{}c@{}}{Depth Map} \\ \cmidrule{2-4} \cmidrule{5-6}
    Surgery phase & PSNR$\uparrow$ & SSIM$\uparrow$ & LPIPS$\downarrow$ & MAE (in cm)$\downarrow$ & Error (in \%)$\downarrow$\\
    \midrule
    Empty OR    & 19.6 & 0.70 & 0.35 & 2.5 & 0.86 \\
    Patient in  & 17.5 & 0.59 & 0.39 & 3.2 & 1.28 \\
    Draping     & 18.4 & 0.62 & 0.43 & 4.3 & 1.76 \\
    Cleaning    & 17.7 & 0.58 & 0.42 & 5.6 & 2.66 \\
    Procedure   & 17.7 & 0.58 & 0.44 & 5.5 & 3.43 \\
    \textbf{Average} & \textbf{18.2} & \textbf{0.61} & \textbf{0.41} & \textbf{4.2} & \textbf{2.00} \\
    \bottomrule
\end{tabular*}
\end{minipage}
\end{center}
\end{table}

\begin{table}[t!]
\begin{center}
\begin{minipage}{\textwidth}
\caption{Ablation of Dynamic DS-NeRF where dynamics and depth supervision are disabled consecutively. The last row gives the performance of a baseline method.}\label{tab:comparison_results}
\begin{tabular*}{\textwidth}{@{\extracolsep{\fill}}lccccc@{\extracolsep{\fill}}}
    \toprule
     & \multicolumn{3}{@{}c@{}}{Colour Image} & \multicolumn{2}{@{}c@{}}{Depth Map} \\ \cmidrule{2-4} \cmidrule{5-6}
    Method & PSNR$\uparrow$ & SSIM$\uparrow$ & LPIPS$\downarrow$ & MAE (in cm)$\downarrow$ & Error (in \%)$\downarrow$\\
    \midrule
    Dynamic DS-NeRF & \textbf{18.2} & \textbf{0.61} & 0.41 & \textbf{4.2} & 2.00 \\
    DS-NeRF & 18.1 & \textbf{0.61} & \textbf{0.39} & 4.3 & \textbf{1.98} \\
    NeRF & 13.9 & 0.45 & 0.59 & 48.4 & 18.91 \\
    \midrule
    Pointcloud (baseline) & 12.0 & 0.41 & 0.59 & 10.8 & 4.38 \\
    \bottomrule
\end{tabular*}
\end{minipage}
\end{center}
\end{table}

\begin{figure}[t!]%
    \centering
    \includegraphics[width=0.8\textwidth]{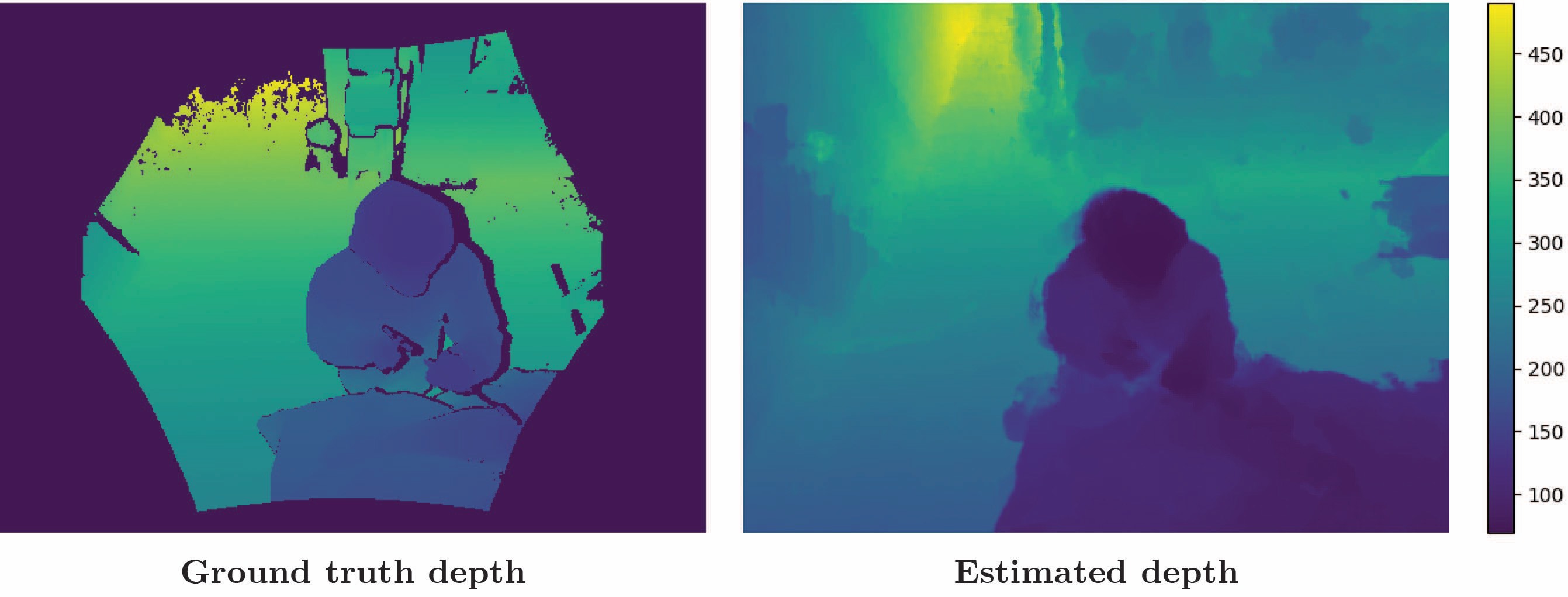}
    \caption{Depth estimation output of Dynamic DS-NeRF for the ``procedure'' phase (right) in comparison with the ground truth depth channel (left). Color bar displays distance in cm.}
    \label{fig:depth_estimation}
\end{figure}

An ablation study is provided in Table~\ref{tab:comparison_results}, where dynamics and depth supervision are disabled consecutively. It can be seen that the dynamic extension does not negatively affect view synthesis performance. The presence of depth supervision has a large positive impact on image quality, with PSNR increasing from 13.9 to 18.1. The depth estimation error decreases drastically from 18.91 to 1.98\%. Figure~\ref{fig:depth_estimation} displays the estimated depth for an unseen camera position in the ``procedure'' phase in comparison with the ground truth depth channel captured from the same camera pose. It can be seen that Dynamic DS-NeRF is able to grasp the geometry of the captured scene accurately. Moreover, the algorithm is able to generate depth values that are not present in the ground truth image due to the depth sensor's hexagon shape or sensor artefacts (e.g., the zero-valued ``shadows''). These results show that depth supervision helps NeRF to reconstruct the scene's 3D geometry accurately, resulting in a higher quality of synthesised images.

\subsection{Dynamic NeRF for Segmentation}
\begin{figure}[t!]%
    \centering
    \includegraphics[width=\textwidth]{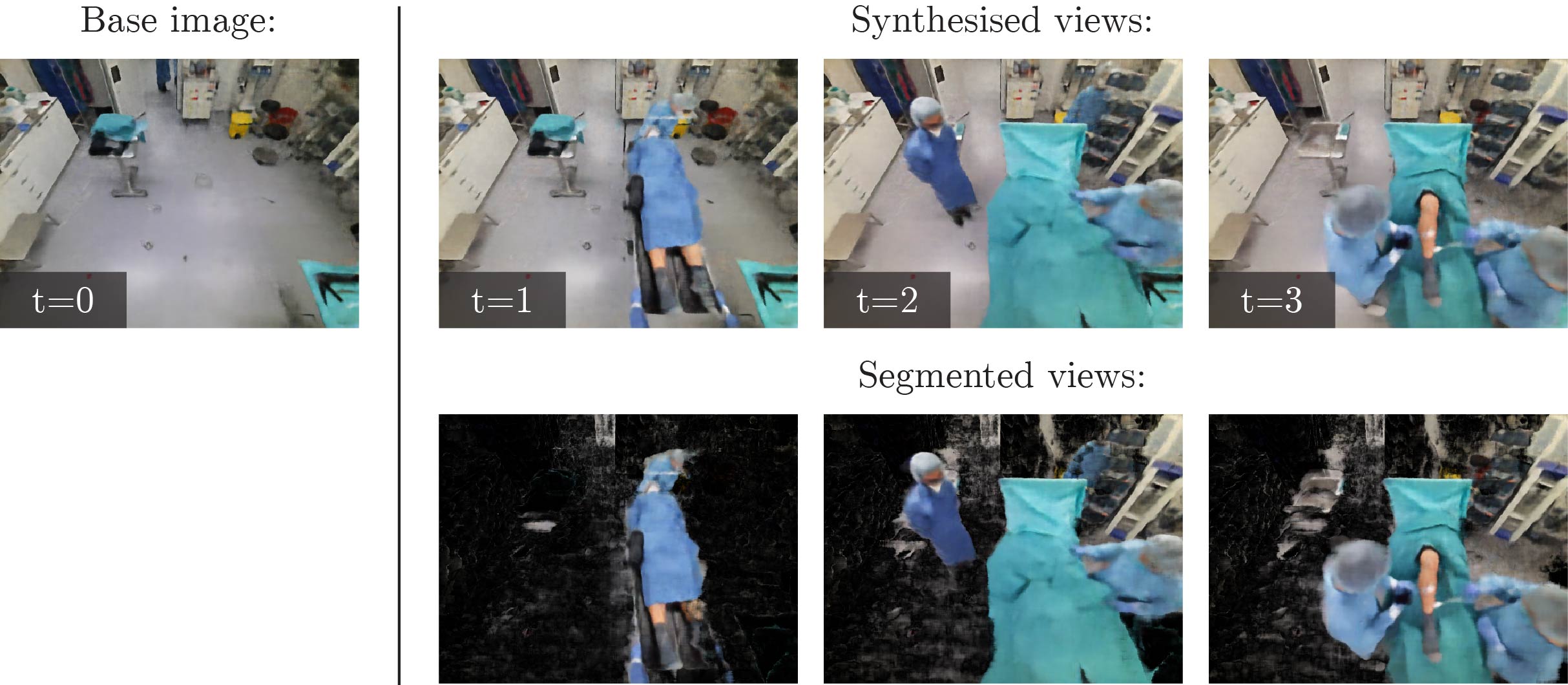}
    \caption{Top row: synthesised views by Dynamic DS-NeRF, where each surgical phase is reconstructed at a specific time interval $t$. Bottom row: segmented views, constructed by subtracting the volume densities at $t=0$ from the densities at $t$.}
    \label{fig:nerf_segmentations}
\end{figure}

Besides the modelling of consecutive video frames, a dynamic NeRF could also represent the scene at distinct moments in time. We use the algorithm to reconstruct an empty OR at $t=0$ and the other phases at $t=\{1, 2, 3, 4\}$. Hence, a single model is optimised to represent these five distinct phases jointly. The top row in Figure~\ref{fig:nerf_segmentations} displays the model output at four time intervals. To demonstrate that Dynamic DS-NeRF could be used for tasks other than view synthesis, we use this model configuration for unsupervised segmentation of objects and people. The representation of NeRF at $t=0$ is considered the base image. This image consists of an empty OR, and we assume that its materials are not of interest as long as they do not move. At this time interval, we sample points along camera rays as usual and store the densities $\sigma^{(0)}$. For the other phases, we sample the densities $\sigma^{(t)}$ of the same point locations at the corresponding time interval. To segment the point locations, we subtract the density of the base image from the density of each other phase:  $\sigma^{(t)} - \sigma^{(0)}$. The resulting densities correspond to material that is present at $t$, but that was not present in the empty OR. We used these resulting densities to render the segmented views. It can be seen that the static floor and left counter disappear from the segmented view, while the surgeons, patient and the anaesthetic tower remain present. Note that we do not simply obtain a difference image, but use the 3D estimated density of the Dynamic DS-NeRF model.

\section{Discussion}
In this work, we explored the use of NeRF for reconstructing surgical scenes in the OR with multi-view RGB-D images from the 4D-OR dataset \cite{ozsoy20224dor}. We showed that the original NeRF does not provide optimal reconstruction results and that the use of depth supervision benefits image quality and removes the necessity to train the algorithm with tens of camera positions, which are difficult to obtain during a surgical procedure. With an additional time variable at the input vector, we showed the possibility of rendering views with a notion of scene dynamics without affecting image quality. When virtually rotating around the surgical field, the synthesised images remain geometrically consistent, even at interpolated camera locations and time values. On the other hand, the images miss fine details and contain artefacts. Training with larger image resolutions or geometric priors can potentially help to produce views with higher quality.

Dynamic DS-NeRF can generate views from an unseen position above the surgical field with 18.2 PSNR, 0.61 SSIM, 0.41 LPIPS and an average depth error of 2.0\%. The method provides better results in comparison to a baseline method that is based on pointcloud projection. Besides a baseline model, we compared three configurations of NeRF: the original NeRF, DS-NeRF and a dynamic DS-NeRF. From the wide range of NeRF architectures available \cite{xie2022neural}, we hypothesise that depth supervision and dynamics are most relevant to the clinical environment. Nevertheless, the exploration of other models, such as pixelNeRF and RegNeRF, would be an interesting follow-up. Other developments in this field are of interest as well, such as methods for speeding up the training of NeRF algorithms \cite{mueller2022instant}.

We envision several potential uses for NeRF in the synthesis of OR images or videos. First, the technology could be used for rendering virtual environments displaying real surgeries in 3D. A spectator could watch a video-recorded procedure while moving the virtual camera to arbitrary camera positions in order to obtain a better view. This could enhance the reviewing or learning process. Second, when combining the 3D reconstruction with virtual reality, one could construct a training exercise for new surgeons or OR staff that shows a photo-realistic instead of simulated environment. This could increase the immersiveness and effectiveness of the training. Third, the implicit neural representations of the NeRF-model can be used for further processing. We showed briefly how these can be used for unsupervised segmentation of relevant objects and people. The resulting pointcloud could help to reduce the number of candidate locations for the detection of objects and human poses. Last, renders from new camera positions could be used to virtually increase the number of cameras in a dataset. This is particularly relevant for OR video datasets, due to limitations for camera placement.

In conclusion, Dynamic DS-NeRF is able to synthesise views of a dynamic surgical field in which the 3D geometry is captured accurately. Depth supervision with RGB-D sensing data increases render quality drastically while requiring fewer camera positions, making the technology applicable to clinical environments.

\section*{Declarations}
This work was sponsored by Johnson \& Johnson MedTech. Jelmer M. Wolterink was supported by NWO domain Applied and Engineering Sciences VENI grant (18192).

%
%
%
%

\bibliographystyle{splncs04}
\bibliography{main}

\begin{thebibliography}{10}
\providecommand{\url}[1]{\texttt{#1}}
\providecommand{\urlprefix}{URL }
\providecommand{\doi}[1]{https://doi.org/#1}

\bibitem{deng2022depth}
Deng, K., Liu, A., Zhu, J.Y., Ramanan, D.: Depth-supervised nerf: Fewer views
  and faster training for free. In: Proc. of the IEEE/CVF Conf. on Comput.
  Vision and Pattern Recognit. pp. 12882--12891 (2022)

\bibitem{gerats20223d}
Gerats, B.G., Wolterink, J.M., Broeders, I.A.: 3d human pose estimation in
  multi-view operating room videos using differentiable camera projections.
  Computer Methods in Biomechanics and Biomedical Engineering: Imaging \&
  Visualization pp.~1--9 (2022)

\bibitem{goldenberg2017using}
Goldenberg, M.G., Jung, J., Grantcharov, T.P.: Using data to enhance
  performance and improve quality and safety in surgery. JAMA surgery
  \textbf{152}(10),  972--973 (2017)

\bibitem{hansen2019}
Hansen, L., Siebert, M., Diesel, J., Heinrich, M.P.: Fusing information from
  multiple 2d depth cameras for 3d human pose estimation in the operating room.
  International journal of computer assisted radiology and surgery
  \textbf{14}(11),  1871--1879 (2019)

\bibitem{li2022neural}
Li, T., Slavcheva, M., Zollhoefer, M., Green, S., Lassner, C., Kim, C.,
  Schmidt, T., Lovegrove, S., Goesele, M., Newcombe, R., et~al.: Neural 3d
  video synthesis from multi-view video. In: Proc. of the IEEE/CVF Conf. on
  Comput. Vision and Pattern Recognit. pp. 5521--5531 (2022)

\bibitem{mildenhall2019local}
Mildenhall, B., Srinivasan, P.P., Ortiz-Cayon, R., Kalantari, N.K.,
  Ramamoorthi, R., Ng, R., Kar, A.: Local light field fusion: Practical view
  synthesis with prescriptive sampling guidelines. ACM Trans. on Graph.
  \textbf{38}(4),  1--14 (2019)

\bibitem{mildenhall2020nerf}
Mildenhall, B., Srinivasan, P.P., Tancik, M., Barron, J.T., Ramamoorthi, R.,
  Ng, R.: Nerf: Representing scenes as neural radiance fields for view
  synthesis. In: ECCV (2020)

\bibitem{mueller2022instant}
M\"uller, T., Evans, A., Schied, C., Keller, A.: Instant neural graphics
  primitives with a multiresolution hash encoding. ACM Trans. on Graph.
  \textbf{41}(4),  1--15 (Jul 2022)

\bibitem{ozsoy20224dor}
\"Ozsoy, E., \"Ornek, E.P., Czempiel, T., Tombari, F., Navab, N.: 4d-or:
  Semantic scene graphs for or domain modeling. In: Int. Conf. on Med. Image
  Comput. and Computer-Assisted Intervention. Springer (2022)

\bibitem{padoy2019}
Padoy, N.: Machine and deep learning for workflow recognition during surgery.
  Minimally Invasive Therapy \& Allied Technologies  \textbf{28}(2),  82--90
  (2019)

\bibitem{pumarola2021d}
Pumarola, A., Corona, E., Pons-Moll, G., Moreno-Noguer, F.: D-nerf: Neural
  radiance fields for dynamic scenes. In: Proc. of the IEEE/CVF Conf. on
  Comput. Vision and Pattern Recognit. pp. 10318--10327 (2021)

\bibitem{rematas2022urban}
Rematas, K., Liu, A., Srinivasan, P.P., Barron, J.T., Tagliasacchi, A.,
  Funkhouser, T., Ferrari, V.: Urban radiance fields. In: Proc. of the IEEE/CVF
  Conf. on Comput. Vision and Pattern Recognit. pp. 12932--12942 (2022)

\bibitem{roessle2022depthpriorsnerf}
Roessle, B., Barron, J.T., Mildenhall, B., Srinivasan, P.P., Nie{\ss}ner, M.:
  Dense depth priors for neural radiance fields from sparse input views. In:
  Proc. of the IEEE/CVF Conf. on Comput. Vision and Pattern Recognit. (June
  2022)

\bibitem{schonberger2016structure}
Schonberger, J.L., Frahm, J.M.: Structure-from-motion revisited. In: Proc. of
  the IEEE Conf. on Comput. Vision and Pattern Recognit. pp. 4104--4113 (2016)

\bibitem{sun2021coil}
Sun, Y., Liu, J., Xie, M., Wohlberg, B., Kamilov, U.S.: Coil: Coordinate-based
  internal learning for tomographic imaging. IEEE Trans. on Comput. Imaging
  \textbf{7},  1400--1412 (2021)

\bibitem{wang2022neural}
Wang, Y., Long, Y., Fan, S.H., Dou, Q.: Neural rendering for stereo 3d
  reconstruction of deformable tissues in robotic surgery. In: Int. Conf. on
  Med. Image Comput. and Computer-Assisted Intervention. pp. 431--441. Springer
  (2022)

\bibitem{wang2004image}
Wang, Z., Bovik, A.C., Sheikh, H.R., Simoncelli, E.P.: Image quality
  assessment: from error visibility to structural similarity. IEEE Trans. on
  Image Process.  \textbf{13}(4),  600--612 (2004)

\bibitem{wei2021nerfingmvs}
Wei, Y., Liu, S., Rao, Y., Zhao, W., Lu, J., Zhou, J.: Nerfingmvs: Guided
  optimization of neural radiance fields for indoor multi-view stereo. In:
  Proc. of the IEEE/CVF Int. Conf. on Comput. Vision. pp. 5610--5619 (2021)

\bibitem{wolterink2022implicit}
Wolterink, J.M., Zwienenberg, J.C., Brune, C.: Implicit neural representations
  for deformable image registration. In: Int. Conf. on Med. Imaging with Deep
  Learning. pp. 1349--1359. PMLR (2022)

\bibitem{xie2022neural}
Xie, Y., Takikawa, T., Saito, S., Litany, O., Yan, S., Khan, N., Tombari, F.,
  Tompkin, J., Sitzmann, V., Sridhar, S.: Neural fields in visual computing and
  beyond. In: Computer Graphics Forum. vol.~41, pp. 641--676. Wiley Online
  Library (2022)

\bibitem{zhang2018unreasonable}
Zhang, R., Isola, P., Efros, A.A., Shechtman, E., Wang, O.: The unreasonable
  effectiveness of deep features as a perceptual metric. In: Proc. of the IEEE
  Conf. on Comput. Vision and Pattern Recognit. pp. 586--595 (2018)

\end{thebibliography}

\pagebreak
\section*{Supplementary materials}

\subsection*{A. Implementation details}
\begin{figure}[h]%
    \centering
    \includegraphics[width=\textwidth]{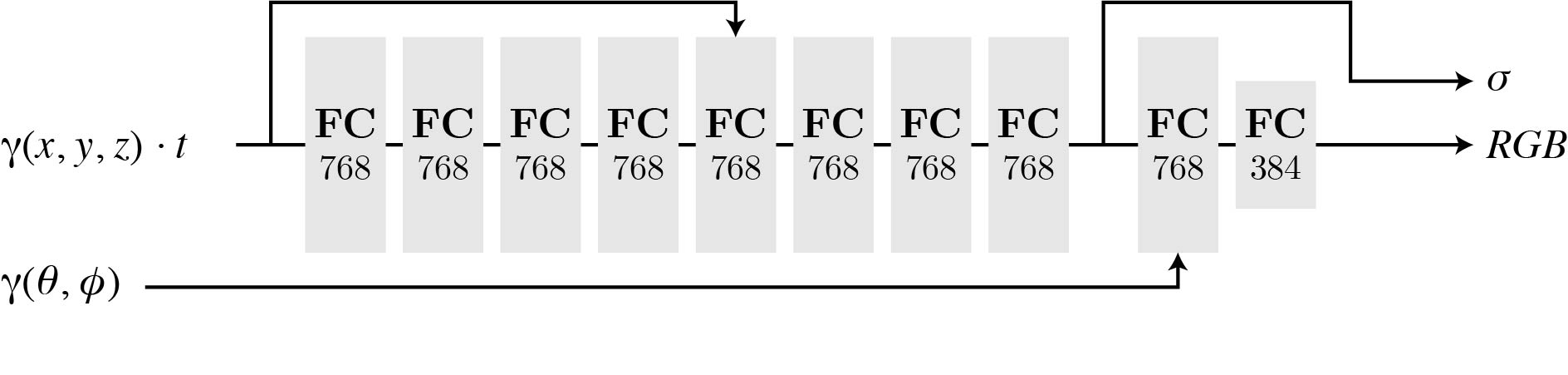}
    \caption{The used NeRF models exists out of eight fully connected (FC) layers that produce material density $\sigma$ from a 3D position $(x, y, z)$ concatenated $(\cdot)$ with time-value $t$. The output of the last FC layer is concatenated with viewing direction $(\theta, \phi)$ and given to two other FC layers that produce colour $RGB$. The 3D position and viewing direction are embedded with positional encoding $\gamma(.)$. The width for all FC layers is 768 neurons except for the last FC layer, which has 384 neurons.}
    \label{fig:nerf_architecture}
\end{figure}

For the implementation of the model, we build upon the depth-supervised NeRF GitHub repository\footnote{https://github.com/dunbar12138/DSNeRF} by Deng \textit{et al.} \cite{deng2022depth}. We adapted the code to read depth values from RGB-D sensor data. Additionally, for each training image we registered a time value $t \in \mathbb{Z}$ that is given to the model for rendering time-dependent scenes. Details of the model are given in Figure~\ref{fig:nerf_architecture}. Note that, similar to the original NeRF \cite{mildenhall2019local}, we use two separate NeRFs for coarse and fine level processing. The architecture and size of these models are equal.

\subsection*{B. Training procedure}
For training the NeRFs, we selected five scenarios in the video and sampled five frames at $t = \{-2, -1, 0, 1, 2\}$ for each scenario. We randomly sampled 100K values in the depth channel of each image, while discarding depth values that are equal to zero. The positions of the cameras used for training were already obtained using external calibration and are given in the dataset. With the RGB-images, depth values, camera poses and time-values we trained a dynamic depth-supervised NeRF for each scenario using the adapted code implementation. Each NeRF is trained to generate images of 512 $\times$ 384 pixels in 50K iterations, with 4096 selected rays per batch, 192 points sampled per ray and a depth loss weighting factor of $0.1$. Other hyper-parameters are equal to the original implementation of depth-supervised NeRF \cite{deng2022depth}.

\subsection*{C. Validation}
For the qualitative validation, we queried the trained NeRFs to synthesise images for a camera path that rotates in 180 degrees around the surgical field. This path exists out of a movement from the right to the middle camera and from the middle to the left camera. Each step $i \in [0, 1, ..., I]$ in this path is an interpolation between two camera matrices:

\begin{equation}
    T_i = \alpha_i T_A + (1 - \alpha_i) T_B,
\end{equation}

where $T_i$ is the rotation-translation matrix of the virtual camera at step $i$, $T_A$ and $T_B$ are the matrices of cameras $A$ and $B$, and $\alpha_i = \frac{i}{I}$ defining the position of the virtual camera at step $i$. To create a video of approximately two seconds with 30 fps, we choose $I = 30$. In the same manner we interpolated between the cameras' focal points. The results shown in Figure~\ref{fig:nerf_results} are at steps $i=0$, $i=15$ and $i=30$.

\end{document}